\documentclass{article}

\usepackage{microtype}
\usepackage{graphicx}
\usepackage{subcaption}
\usepackage{booktabs} 

\usepackage{hyperref}

\usepackage[accepted]{icml2026}

\makeatletter
\renewcommand{\Notice@String}{%
\textit{Accepted at the ICML 2026 Workshop on Forecasting as a New Frontier of Intelligence},
Seoul, South Korea, 2026.}
\makeatother

\usepackage{amsmath}
\usepackage{amssymb}
\usepackage{mathtools}
\usepackage{amsthm}

\usepackage[capitalize,noabbrev]{cleveref}

\usepackage{placeins}

\theoremstyle{plain}

\theoremstyle{definition}

\theoremstyle{remark}

\usepackage[textsize=tiny]{todonotes}

\newcommand{\model}[1]{\textit{#1}}

\icmltitlerunning{Diversity is the Strength of the AI Crowd}

\begin{document}

\twocolumn[
  \icmltitle{Diversity is the Strength of the AI Crowd}



  \icmlsetsymbol{equal}{*}

  \begin{icmlauthorlist}
    \icmlauthor{Matthew Aitchison}{mantic}
    \icmlauthor{Scott Jeen}{mantic}
    \icmlauthor{Toby Shevlane}{mantic}
    \icmlauthor{Ben Day}{mantic}
  \end{icmlauthorlist}

  \icmlaffiliation{mantic}{Mantic Technologies, London, UK}
  \icmlcorrespondingauthor{Matthew Aitchison}{matthew@mantic.com}

  \icmlkeywords{Machine Learning, ICML}

  \vskip 0.3in
]



\printAffiliationsAndNotice{}  

\begin{abstract}
Top AI forecasting systems are approaching superforecaster-level accuracy on future world events, but still rely primarily on off-the-shelf LLMs combined with forecasting-specific context gathering and scaffolding. We study how to improve this recipe through ensembling: given a fixed number of samples, which off-the-shelf model forecasts should be combined to maximize accuracy? On binary questions from the Metaculus AI Benchmark, we find that individual accuracy is not enough: many frontier LLMs make highly correlated predictions, limiting the value of additional forecasts from the same or similar models. Instead, the strongest ensembles combine accurate but diverse forecasters, with models such as \model{Grok 4} contributing disproportionately because their predictions are less correlated with other frontier LLMs. These results suggest that the strength of the AI crowd comes not from sampling more forecasts indiscriminately, but from combining forecasts across models with complementary errors, motivating forecasting systems that explicitly optimize for both model quality and diversity.
\end{abstract}

\section{Introduction}\label{sec:introduction}

Top-performing AI forecasting systems are now approaching skilled human forecasters on tournament-style questions about future world events~\citep{halawi2024approaching, schoenegger2024wisdom}.

The prevailing recipe in current forecasting tournaments wraps an off-the-shelf large language model in a forecasting-specific pipeline that handles, for example, retrieval, structured prompting, and multi-sample aggregation. Practitioners building such systems face a cost/time constraint the literature rarely addresses head-on: given a fixed number of forecasts per question, which model forecasts should be combined to maximise performance?

A natural starting place would be to draw all samples from the strongest single model. We show this answer is wrong, for a direct reason: samples from well-performing (and indeed less well-performing) frontier LLMs are highly correlated with each other\footnote{Correlation here is between the predicted probabilities; the reasoning traces that produce them may differ.}. A second draw from a single model, whose distribution closely tracks the first, contributes only a fraction of the information that one draw from a similarly skilled but more diverse model would contribute. Classical ensemble theory makes this precise: an averaging ensemble's error decomposes into a member-accuracy term and a diversity term~\citep{krogh1994neural,wood2023unified}, so the strength of a multi-LLM crowd comes from sampling models with \emph{complementary} errors, not from sampling more indiscriminately.

We evaluate this claim on binary questions from the Metaculus AI Benchmark Q2 2025 tournament, with five forecasters representing contemporary frontier LLMs and one fine-tuned variant of \model{gpt-oss-120b}. We brute-force-evaluate every valid sample allocation up to budget $B{=}5$, and report three findings. First, the strongest off-the-shelf frontier LLMs make tightly correlated predictions; \model{Grok 4} and the fine-tuned model sit further from this cluster on a pairwise Jensen--Shannon-divergence diversity matrix. Second, the optimal fixed-budget allocation never concentrates all samples on a single model: at $B{=}5$ it splits the budget across the fine-tuned model, \model{Gemini 3 Pro}, \model{GPT-5}, and \model{Grok 4}. Third, \model{Grok 4} is the \emph{least-replaceable} member of the optimum despite ranking third in solo accuracy: its outsize contribution to the ensemble comes from its low correlation with the rest of the pool, not from its solo skill. \footnote{An earlier, non-archival version of these findings appeared in a \href{https://thinkingmachines.ai/news/training-llms-to-predict-world-events/}{technical blog post}.}

\begin{figure}[!t]
  \centering
  \includegraphics[width=\columnwidth]{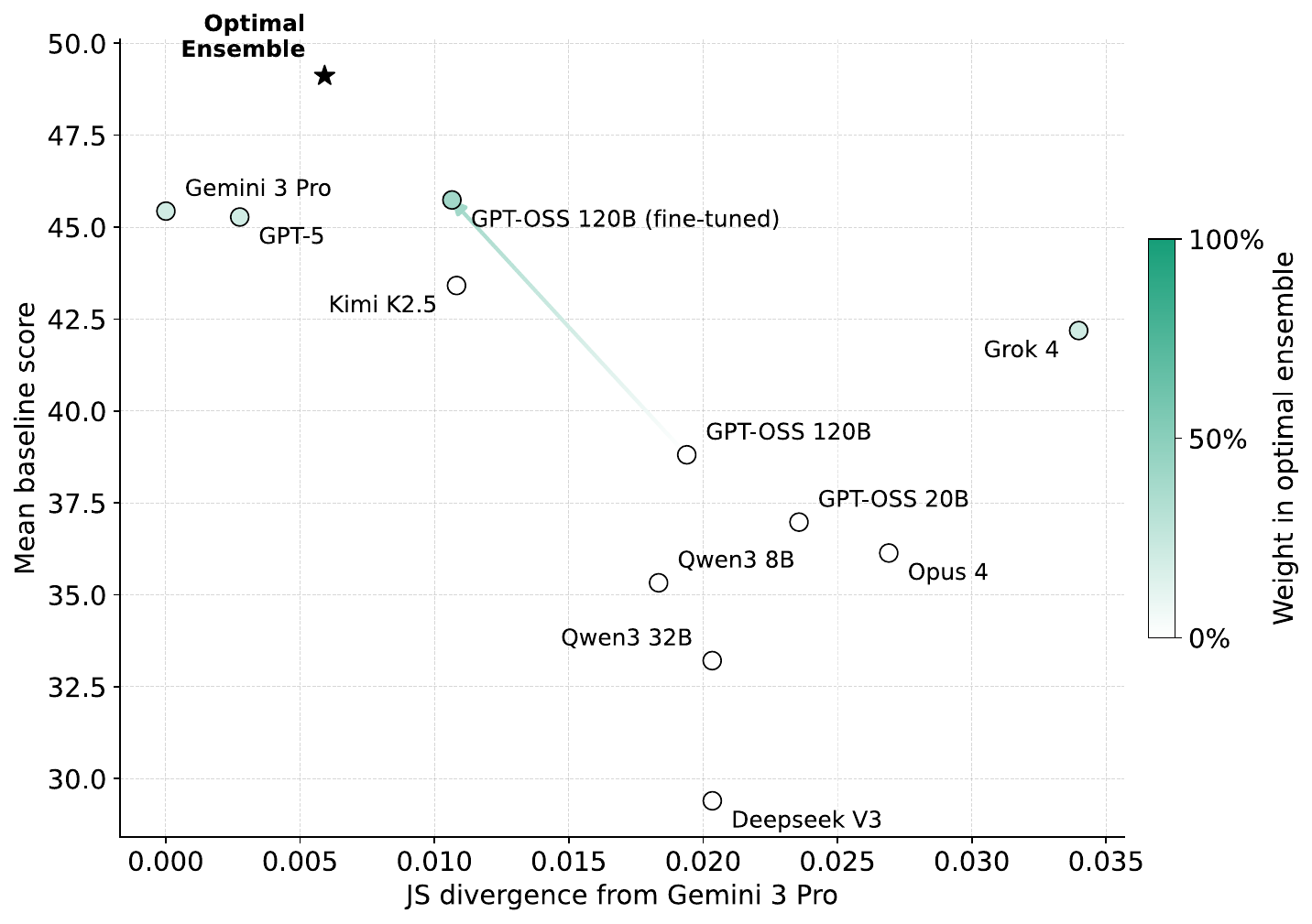}
  \caption{\textbf{Diversity pays off.} Mean baseline score on binary
  event questions from the Metaculus AI Benchmark Q2 2025 against the
  test-averaged Jensen--Shannon divergence from \model{Gemini 3 Pro}
  for our five evaluated models. Marker shade indicates each model's
  weight in the optimal $B{=}5$ ensemble $a^\star_5 =
  (\mathrm{FT}{:}\,2, \mathrm{Gem}{:}\,1, \mathrm{GPT}{:}\,1,
  \mathrm{Grok}{:}\,1, \mathrm{Kimi}{:}\,0)$. See \S\ref{sec:res-corr}
  for discussion.}
  \label{fig:pairwise}
\end{figure}

\section{Related Work}\label{sec:related-work}

\paragraph{AI forecasting.}
Top AI forecasting systems now approach skilled human forecasters on tournament-style questions~\citep{halawi2024approaching, schoenegger2024wisdom}. The standard recipe is shared: an off-the-shelf LLM wrapped in retrieval-augmented reasoning, with multi-sample aggregation, evaluated on dynamic benchmarks such as ForecastBench~\citep{karger2025forecastbench} and FutureX~\citep{zeng2025futurex} that draw resolution-unknown questions to avoid data leakage. Two axes of improvement have been pursued. The first strengthens the per-question pipeline via retrieval-augmented reasoning~\citep{halawi2024approaching}, structured belief updating~\citep{murphy2026agentic}, or task-specific fine-tuning~\citep{turtel2025llms, turtel2025outcome}. The second ensembles across multiple LLMs: aggregating 12 frontier models matches a tournament of $\sim$900 human forecasters~\citep{schoenegger2024wisdom}. Within this second axis, prior work has implicitly assumed uniform allocation across models; we instead ask which non-uniform allocation maximises ensemble accuracy under a fixed sample budget.

\paragraph{Ensemble theory and LLM ensembling.}
Classical ensemble learning gives a clean characterisation of when ensembling helps: under squared loss the error of an averaging ensemble decomposes as \emph{average member error minus a non-negative diversity term}~\citep{krogh1994neural,brown2005diversity}; the same decomposition extends to broader loss families, including the log-score we use, via the recent unified theory of~\citet{wood2023unified}. The implication that ensembles benefit from \emph{both} accurate \emph{and} decorrelated members is old; what this paper adds is an empirical characterisation of the trade-off in an LLM-crowd setting, where members are heterogeneous frontier LLMs rather than resamples of a single base learner. Existing LLM-ensemble work focuses on text-generation outputs, using ranking-based or generative fusion~\citep{jiang2023llmblender}, or studies weight-space averaging within a model family~\citep{wortsman2022soups}. We instead ensemble probabilistic forecasts and quantify how much of an ensemble's gain comes from per-model accuracy versus per-pair diversity.

\section{Preliminaries}\label{sec:preliminaries}

\paragraph{Forecasting questions.}
A \emph{binary forecasting question} $q$ is a statement whose truth is
revealed at a known resolution time, paired with a ground-truth outcome
$y\in\{0,1\}$. We write a test set as $\mathcal{D}=\{(q_i, y_i)\}_{i=1}^N$.

\paragraph{Forecasters and ensembles.}
A \emph{forecaster} $\pi_m$ for model $m\in\mathcal{M}$ is the full prediction pipeline: a model wrapped in the scaffolding described in \S\ref{sec:method}, not the raw LLM call. Each forecaster is a stochastic map from a question to a probability of the affirmative outcome; $\hat{p}_{m,i}^{(r)} \in [0,1]$ denotes the $r$-th sample drawn from $\pi_m$ on question $q_i$. Sample-level stochasticity comes from nonzero sampling temperature, retrieval variability, and inference-kernel non-determinism that cannot generally be switched off~\citep{thinking_machines_nondeterminism}; we treat all such variability as part of $\pi_m$. We write $\bar{p}_{m,i} = \frac{1}{R_m}\sum_{r=1}^{R_m} \hat{p}_{m,i}^{(r)}$ for the per-question average over $R_m$ independent samples. A \emph{weighted ensemble} of $M$ models with weights $\omega = (\omega_1,\dots,\omega_M)\in\Delta^{M-1}$ predicts
\begin{equation}
  \hat{p}^{\,\omega}_i \;=\; \sum_{m=1}^{M} \omega_m \, \bar{p}_{m,i},
  \label{eq:ensemble-weighted}
\end{equation}
followed by a small clipping operation that bounds each side away from $0$ and $1$ by a fixed mass $\varepsilon = 0.05$; see Appendix~\ref{sec:appendix-aggregation} for the operator details. We treat this aggregation operator as fixed throughout: the question we study is which models contribute the weight, not how the contributions are combined. A \emph{discrete team} of total sample budget $B$ is the special case of integer weights $\omega_m = c_m / B$ with $c_m \in \mathbb{Z}_{\ge 0}$ and $\sum_m c_m = B$, which we use in \S\ref{sec:method} to define replaceability.

\paragraph{Scoring and diversity.}
We evaluate ensembles with the Metaculus \emph{baseline score}, a rescaled log score under which a uniform $50\%$ prediction earns $0$ points and perfect foresight earns $100$~\citep{metaculus_baseline_score}; the score is strictly proper~\citep{gneiting2007strictly}. We report per-question baseline points (BP) and mean BP over the test set. To quantify how much two forecasters disagree on a question, we use the symmetric Jensen--Shannon (JS) divergence between the two Bernoulli predictive distributions induced by their probabilities. Writing $\bar{p} = \tfrac{1}{2}(\hat{p}_a + \hat{p}_b)$ for the mixture distribution, the JS divergence is
\begin{equation}
  \mathrm{JS}\!\left(\hat{p}_a,\hat{p}_b\right) \;=\; \tfrac{1}{2}\,\mathrm{KL}\!\left(\hat{p}_a \,\Vert\, \bar{p}\right) + \tfrac{1}{2}\,\mathrm{KL}\!\left(\hat{p}_b \,\Vert\, \bar{p}\right).
  \label{eq:js}
\end{equation}
We average JS across the test set to give a per-pair diversity score $\mathrm{JS}(m,m')$; for a model $m$ relative to a set $S$ we write $\mathrm{JS}(m, S) = \frac{1}{|S|}\sum_{m'\in S}\mathrm{JS}(m,m')$.

\section{Methods}\label{sec:method}

\paragraph{Models.}
We evaluate four off-the-shelf frontier LLMs (\model{Gemini 3 Pro}, \model{GPT-5}, \model{Grok 4}, and \model{Kimi K2.5}) and one fine-tuned model, \model{FT-gpt-oss-120b}: a fine-tuned \model{gpt-oss-120b} trained with reinforcement learning on a training set of forecasting questions disjoint from our test set. Each model is wrapped in the same two-stage scaffolding pipeline: a retrieval phase that gathers question-specific evidence, followed by a structured prediction phase in which the model emits a probability for the binary question. We use each model's default sampling temperature. Between-run variation comes from sampling stochasticity and inference-kernel non-determinism (\S\ref{sec:preliminaries}). We draw $R_m=3$ independent samples per model per question.

\paragraph{Data.}
Our test set is the binary subset of the Metaculus AI Benchmark Q2 2025 tournament, $N{=}113$ binary event questions resolved between May and July 2025.\footnote{\href{https://gist.github.com/enjeeneer/86e24a52e6041a3d78e333bcab16984d}{Full list of questions}.}
We run the research phase upfront to collect information that would have been available at prediction time, and pair the output with the question to produce static prompts. The knowledge cutoff of every evaluated model precedes the tournament's start, and retrieval is limited to sources available before each question's resolution date. A sample of questions is given in Appendix~\ref{sec:appendix-dataset}.

\begin{figure*}[!tb]
  \centering
  \includegraphics[width=1.0\textwidth]{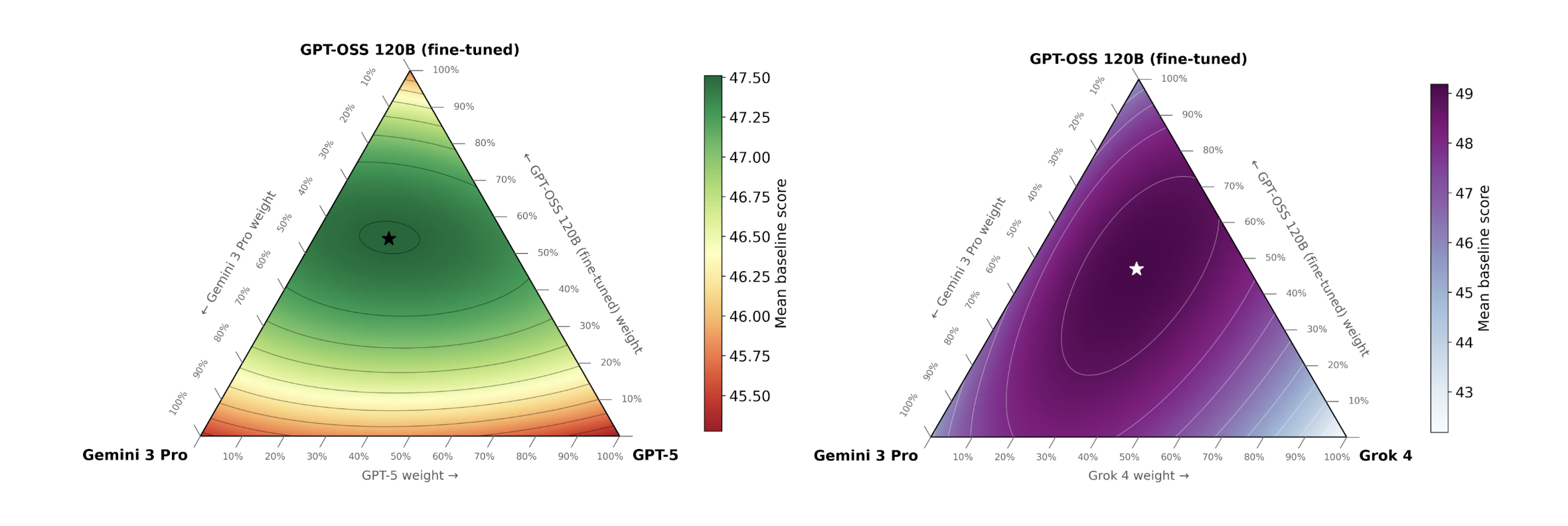}
  \caption{\textbf{Three-way ensembles.} Mean baseline score over the
  three-model weighted ensemble simplex. \emph{Left:}
  \model{FT-gpt-oss-120b}, \model{Gemini 3 Pro}, and \model{GPT-5};
  the optimum ($\star$) lies at $(\omega_{\mathrm{FT}},
  \omega_{\mathrm{Gem}}, \omega_{\mathrm{GPT}}) \approx
  (0.56, 0.26, 0.18)$. \emph{Right:} \model{Grok 4} substitutes for
  \model{GPT-5}; the optimum lies at $(\omega_{\mathrm{FT}},
  \omega_{\mathrm{Gem}}, \omega_{\mathrm{Grok}}) \approx
  (0.48, 0.26, 0.26)$. In both, the fine-tuned model receives about
  half the weight.}
  \label{fig:ternary}
\end{figure*}

\paragraph{Continuous-weight sweeps.}
We characterise the ensemble in two complementary ways. For Figure~\ref{fig:pairwise} we report each model's solo baseline score against its test-averaged Jensen--Shannon divergence from \model{Gemini 3 Pro}, with marker shade encoding its weight in the discrete-team optimum $a^\star_5$ (defined below). For Figure~\ref{fig:ternary} we grid the 2-simplex of three-model mixing weights with step $0.05$ and score each $(\omega_1, \omega_2, \omega_3)$ point under Equation~\ref{eq:ensemble-weighted} on the test set.

\paragraph{Replaceability.}
At $B{=}5$ over the five-model pool with per-model cap $c_m \le 3$, exactly $101$ allocations are valid (out of $126$ unconstrained). We enumerate all of them, score each with Equation~\ref{eq:ensemble-weighted} under the integer weights $\omega_m = c_m / B$, and identify the optimal team $a^\star_5 = \arg\max_{a} S(a)$, where $S(a) = \frac{1}{N}\sum_i \mathrm{bs}(\hat{p}^{\,a}_i, y_i)$ is the mean baseline score. The \emph{replaceability} of a model $m$ in $a^\star_5$ is the score gap to the best team that \emph{excludes} $m$,
\begin{equation}
  \Delta_m \;=\; S(a^\star_5) \;-\; \max_{a:\, c_m = 0} S(a),
  \label{eq:replaceability}
\end{equation}
where the maximum is over the same allocation universe with the additional constraint $c_m = 0$. A large positive $\Delta_m$ identifies $m$ as \emph{least-replaceable}: dropping it from the optimum costs the most relative to the best alternative team. By construction $\Delta_m = 0$ for any model not in $a^\star_5$.

\section{Results}\label{sec:results}

\subsection{Pairwise mixing with a frontier anchor}\label{sec:res-corr}

Figure~\ref{fig:pairwise} plots each model's mean baseline score against
its test-averaged Jensen--Shannon divergence from \model{Gemini 3 Pro},
with marker shade encoding the model's weight in the optimal $B{=}5$
team $a^\star_5$. Most frontier LLMs (\model{GPT-5}, \model{Kimi K2.5})
cluster at low JS divergence from \model{Gemini 3 Pro}, scoring well
but contributing little diversity. \model{Grok 4} and
\model{FT-gpt-oss-120b} sit further out on the JS axis, scoring
comparably with the frontier cluster while correlating less, and
together with \model{Gemini 3 Pro} and \model{GPT-5} they make up the
four models that earn a slot in $a^\star_5$.

\subsection{Three-way mixing}\label{sec:res-allocation}

The pairwise sweep tracks each model against a single anchor and misses
joint effects. Figure~\ref{fig:ternary} (left) reports the three-way
heatmap over \model{GPT-5}, \model{Gemini 3 Pro}, and
\model{FT-gpt-oss-120b}, gridded across the 2-simplex with step $0.05$.
The optimum lies near $(\omega_{\mathrm{FT}}, \omega_{\mathrm{Gem}},
\omega_{\mathrm{GPT}}) = (0.56,\,0.26,\,0.18)$: the fine-tuned model
receives the largest weight. The heatmap is asymmetric: moving along the
\model{GPT-5}--\model{Gemini 3 Pro} edge (no FT) gains at most $\approx
0.6$~BP over the better single-model corner, whereas moving toward the
FT vertex picks up $\approx 2.7$~BP. The right panel of
Figure~\ref{fig:ternary} substitutes \model{Grok 4} for \model{GPT-5}
and shows the same qualitative asymmetry.

\begin{figure}[!t]
  \centering
  \includegraphics[width=\columnwidth]{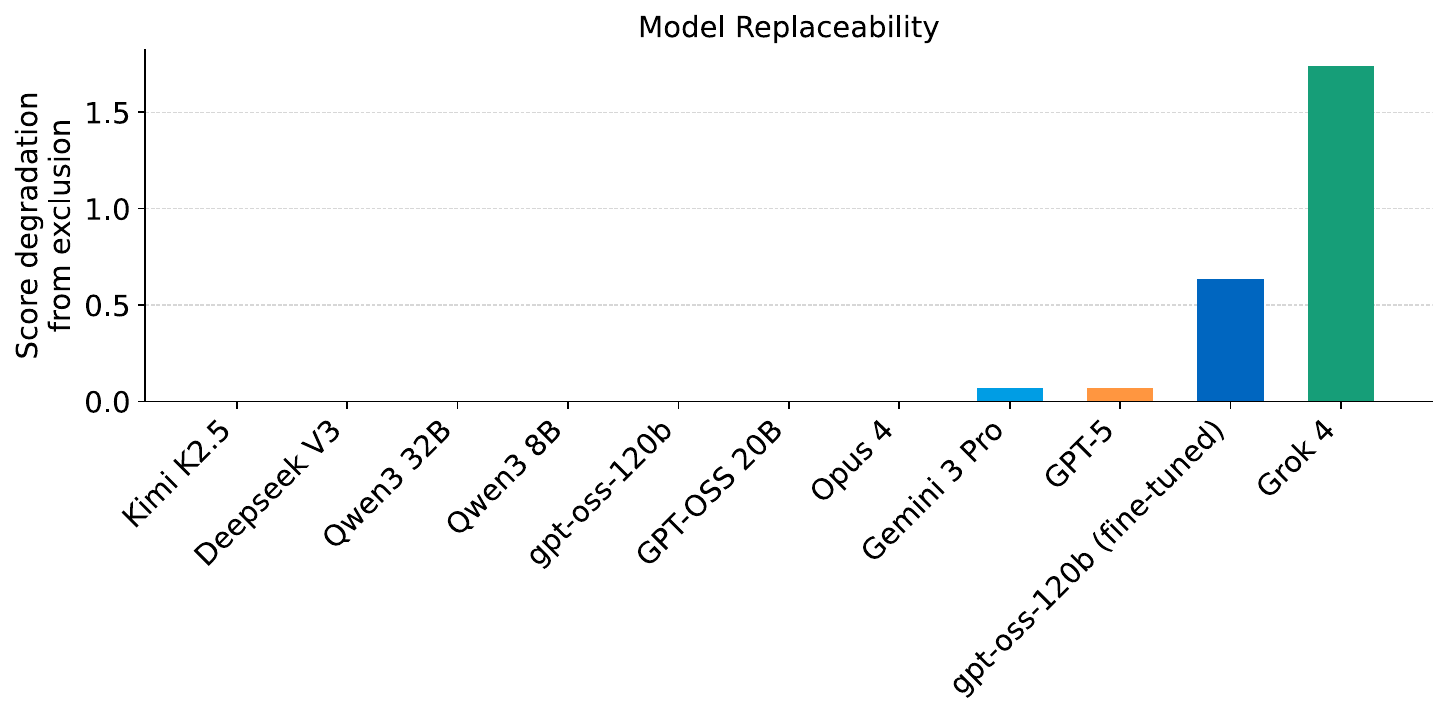}
  \caption{\textbf{When selecting an ensemble of frontier and
  open-source models, \model{Grok 4} and fine-tuned
  \model{gpt-oss-120b} are the least replaceable.} Model
  replaceability is defined as the reduction in score incurred when
  removing a model from the optimal ensemble
  (Eq.~\ref{eq:replaceability}). By definition, if a model is not
  included in the optimal ensemble, there is no cost to removing it.}
  \label{fig:replaceability}
\end{figure}

\subsection{Replaceability under a $B{=}5$ budget}\label{sec:res-replaceability}

Switching from continuous weights to integer sample budgets, we
brute-force every $B{=}5$ allocation over the five-model pool (101
valid allocations under the per-model cap $c_m\le 3$). The optimal
team is $a^\star_5 = (\mathrm{FT}{:}\,2,\,\mathrm{Gem}{:}\,1,\,
\mathrm{GPT}{:}\,1,\,\mathrm{Grok}{:}\,1)$.
Figure~\ref{fig:replaceability} reports the replaceability
$\Delta_m$ (Equation~\ref{eq:replaceability}) for each member of
$a^\star_5$: the score gap from $a^\star_5$ to the best team that
\emph{excludes} $m$. \model{Grok 4} is the largest contributor by a
wide margin ($\Delta_{\mathrm{Grok}} \approx 1.7$~BP), with the
fine-tuned model second ($\Delta_{\mathrm{FT}} \approx 0.6$~BP).
\model{GPT-5} and \model{Gemini 3 Pro} contribute $<0.1$~BP each,
indicating that the two frontier-cluster members are largely
interchangeable within the optimal team.
\model{Grok 4} ranks only third in solo baseline score, so its outsize
ensemble contribution comes from its low correlation with the frontier
cluster (\S\ref{sec:res-corr}), which is exactly the quantity classical
ensemble theory predicts should govern the diversity term of an
ensemble's error decomposition~\citep{krogh1994neural,wood2023unified}.
We make this connection explicit in
Appendix~\ref{sec:appendix-decomposition}.

Absent bootstrap confidence intervals, we treat the contribution ranking as the robust result and the exact magnitudes as indicative.

\section{Discussion}\label{sec:discussion}

The pattern across our three results is consistent with the diversity term in the classical ambiguity decomposition~\citep{krogh1994neural,wood2023unified}: at a fixed sample budget the marginal value of an additional forecast decomposes into a solo-accuracy contribution and a diversity contribution. In our pool the five solo scores cluster within a few BP of each other (\S\ref{sec:res-corr}), so it is the diversity contribution that differentiates additions, and the most-decorrelated model in the pool, \model{Grok 4}, is the least-replaceable member of the optimal team (\S\ref{sec:res-replaceability}) despite ranking only third in solo accuracy. For practitioners building forecasting systems, this means the natural strategy of stacking samples from the strongest single model is suboptimal relative to mixing in a decorrelated alternative.

For model developers the implication is more speculative but worth flagging: training pipelines that incidentally decorrelate via distinct pre-training (\model{Grok 4} is plausibly one such case in our pool) or deliberately via task-specific fine-tuning (\model{FT-gpt-oss-120b}) carry ensemble-level value that is not visible in a head-to-head solo accuracy comparison. A model that is third in solo accuracy can still be first in marginal ensemble contribution.

\section{Limitations}\label{sec:limitations}

Our study uses $N{=}113$ binary event questions from a single tournament, caps independent runs per model at $R_m=3$, ignores per-sample cost differences across models, and uses one replaceability definition (leave-one-out re-optimisation) among several plausible choices.

\section{Conclusion}\label{sec:conclusion}

Across pairwise, three-way, and discrete-team analyses on the Metaculus AI Benchmark Q2 2025 binary set, we have shown that the strength of the AI crowd lies in combining forecasters that are both accurate and decorrelated.

\section*{Impact Statement}

This paper studies how to ensemble off-the-shelf language-model forecasters. Improvements in automated forecasting could inform decision-making across the economy and in government~\citep{tetlock2015superforecasting,karger2025forecastbench}; the specific contribution here, identifying which models contribute most to a forecasting ensemble, is intended to make existing forecasting systems more accurate, not to change the kinds of decisions those systems are used for. We see no specific ethical or societal harms beyond those already discussed in the broader AI forecasting literature.

\bibliography{diversity}
\bibliographystyle{icml2026}

\newpage
\appendix
\onecolumn

\section{Ambiguity decomposition of the ensemble's gain}\label{sec:appendix-decomposition}

We make explicit the connection between the empirical replaceability
ranking of Section~\ref{sec:res-replaceability} and the \emph{ambiguity
decomposition} of an averaging ensemble. For a single binary question
with outcome $y\in\{0,1\}$, member predictions $\hat{p}_1,\dots,\hat{p}_B$,
and uniform ensemble $\bar{p}=\tfrac{1}{B}\sum_b \hat{p}_b$, Krogh and
Vedelsby's identity~\citep{krogh1994neural} gives the squared error of
the ensemble as the average squared error of its members minus a
non-negative diversity term,
\begin{equation}
  (\bar{p}-y)^2 \;=\; \frac{1}{B}\sum_{b=1}^{B}(\hat{p}_b-y)^2
  \;-\; \frac{1}{B}\sum_{b=1}^{B}(\hat{p}_b-\bar{p})^2.
  \label{eq:ambiguity}
\end{equation}
This identity is exact under squared loss. \citet{wood2023unified}
generalise it to arbitrary Bregman divergences, including the log-score
behind the Metaculus baseline metric we report: the ensemble strictly
improves on the average member exactly to the extent that members
disagree, regardless of the choice of Bregman loss. The
$\Delta_m$ ranking of Section~\ref{sec:res-replaceability} is the
empirical counterpart of the per-member diversity term: a model whose
removal forces the next-best team to be substantially worse is one
that contributed disproportionately to the diversity term of the
ensemble it was part of.

\section{Aggregation operator and haircut parameters}\label{sec:appendix-aggregation}

The ensemble probability $\hat{p}^{\,\omega}_i$ used throughout
(Eq.~\ref{eq:ensemble-weighted}) is followed by a small clipping
operation that bounds the affirmative-outcome probability away from
$\{0,1\}$. Concretely, the raw weighted mean $\bar{p}\in[0,1]$ is
replaced by
\begin{equation*}
  \hat{p} \;=\; \max\!\bigl(\varepsilon,\,
                \min(1-\varepsilon,\,\bar{p})\bigr),
\end{equation*}
where $\varepsilon = 0.05$ is the per-side clipping mass. This matches
the binary special case of the projection used inside the forecasting
pipeline that supplied our predictions and is required for the
log-score to be finite on resolution.

All results reported in the main body are under this fixed aggregation
operator; results under the unclipped mean and under larger
$\varepsilon$ are qualitatively unchanged but shift mean BP by a small
additive constant.

\section{Question sample}\label{sec:appendix-dataset}

To give the reader a sense of the kinds of questions in the test set
(their topics, framings, and resolution timing), we list three
examples drawn from the $N{=}113$ binary event questions of
the Metaculus AI Benchmark Q2 2025 tournament.
\begin{itemize}
  \item \emph{Will Haley Stevens announce her candidacy for US Senator
        from Michigan before May 1, 2025?} Resolution date 2025-05-01.
        Resolved: yes.
  \item \emph{Before July 1, 2025, will the government of Greenland
        officially announce a date for an independence referendum?}
        Resolution date 2025-07-01. Resolved: no.
  \item \emph{Before July 1, 2025, will Discord announce that it is
        planning an IPO?} Resolution date 2025-07-01. Resolved: no.
\end{itemize}


\end{document}